\documentclass{article}
\usepackage{ijcai19}

\usepackage{times}
\usepackage{soul}
\usepackage{url}
\usepackage[hidelinks]{hyperref}
\usepackage[utf8]{inputenc}
\usepackage[small]{caption}
\usepackage{graphicx}
\usepackage{amsmath}
\usepackage{booktabs}
\usepackage{amsthm}
\usepackage{amsfonts}
\usepackage[T1]{fontenc}

\usepackage{quoting}

\usepackage[multiple]{footmisc}

\usepackage{fancybox}
\usepackage{amsfonts}

\usepackage{environ}
\usepackage{mathnotation}
\usepackage{multirow}

\usepackage{color}

\usepackage{xspace}
\usepackage{dashrule}
\usepackage[table,xcdraw]{xcolor}
\usepackage[inline]{enumitem}

\newcommand{\dpi}{\mathit{dpi}}
\newcommand{\sol}{\mathbf{diag}}
\newcommand{\conf}{\mathbf{conf}}
\newcommand{\dt}{\mathcal{D}^*}
\newcommand{\md}{\mathcal{D}}
\newcommand{\mD}{\mathbf{D}}
\newcommand{\mC}{\mathbf{C}}

\newcommand{\qqm}{\mathsf{heur}}
\newcommand{\state}{\mathsf{state}}
\newcommand{\nd}{\mathsf{nd}}
\newcommand{\node}{\mathsf{n}}
\newcommand{\ld}{\mathit{ld}}
\newcommand{\mo}{\mathcal{K}}
\newcommand{\mb}{\mathcal{B}}
\newcommand{\tax}{\mathit{ax}}
\newcommand{\Tp}{\mathit{P}}
\newcommand{\Tn}{\mathit{N}}
\newcommand{\tp}{\mathit{p}}
\newcommand{\tn}{\mathit{n}}
\newcommand{\mc}{\mathcal{C}}

\newcommand{\Queue}{{\mathbf{Q}}}

\newcommand{\pr}{\mathit{pr}}

\newcommand{\XSup}[1]{\times_{(\supset #1)}}

\usepackage{algorithm}
\usepackage[noend]{algpseudocode}
\algrenewcommand\algorithmicrequire{\textbf{Input:}}
\algrenewcommand\algorithmicensure{\textbf{Output:}}
\algrenewcommand\alglinenumber[1]{\tiny #1:} 
\algnewcommand{\IfThen}[2]{
	\State \algorithmicif\ #1\ \algorithmicthen\ #2
}

\usepackage[all]{xy}
\usepackage{tikz}

\newcounter{examplecounter}
\newenvironment{example}{
	\refstepcounter{examplecounter}%
	
	\vspace{7pt}
	\noindent\textbf{Example \arabic{examplecounter}}%
	\quad
}{
	
	\vspace{7pt}
}

\newtheorem{theorem}{Theorem}

\newtheorem{property}{Property}

\begin{document}
\title{
Towards Optimizing Reiter's HS-Tree for Sequential Diagnosis}
%
%
%
%
%
%
%
%

\author{
	Patrick Rodler
	\affiliations
	Alpen-Adria Universit\"at Klagenfurt, 9020 Klagenfurt, Austria\\
	\emails
	patrick.rodler@aau.at
}

\maketitle              
\begin{abstract}
	Reiter's HS-Tree is one of the most popular diagnostic search algorithms due to its desirable properties and
	general applicability. In sequential diagnosis, where the addressed diagnosis problem is subject to successive change through the acquisition of additional knowledge about the diagnosed system, HS-Tree is used in a stateless fashion. That is, the existing search tree is discarded when new knowledge is obtained, albeit often large parts of the tree are still relevant and have to be rebuilt in the next iteration, involving redundant operations and costly reasoner calls. 	As a remedy to this, we propose DynamicHS, a variant of HS-Tree that avoids these redundancy issues by maintaining state throughout sequential diagnosis while preserving all desirable properties of HS-Tree. Preliminary results of ongoing evaluations in a problem domain where HS-Tree is the state-of-the-art diagnostic method suggest significant time savings achieved by DynamicHS by reducing expensive reasoner calls.
\end{abstract}
\section{Introduction}
\label{sec:intro}
In model-based diagnosis, given a \emph{diagnosis problem instance (DPI)}---consisting of the \emph{system description}, the system \emph{components}, and the \emph{observations}---a \emph{(minimal) diagnosis} is an (irreducible) set of components that, when assumed abnormal, 
leads to consistency between system description (predicted behavior) and observations (real behavior).
In many cases, there 
is a substantial number
of competing (minimal) diagnoses. 
To isolate the \emph{actual diagnosis} (which pinpoints the \emph{actually} faulty components), \emph{sequential diagnosis} \cite{dekleer1987} methods collect additional system observations (called \emph{measurements}) to gradually refine the set of diagnoses.

One of the most popular and widely used algorithm for the computation of 
diagnoses
in model-based diagnosis is Reiter's HS-Tree \cite{Reiter87}. It is adopted in various domains such as for the debugging of software \cite{abreu2011simultaneous,wotawa2010fault} or ontologies and knowledge bases \cite{friedrich2005gdm,Rodler2015phd,Kalyanpur2006a,meilicke2011}, or for the diagnosis of hardware \cite{friedrich1999model},
recommender systems \cite{felfernig2008intelligent}, configuration systems \cite{DBLP:journals/ai/FelfernigFJS04},
and circuits \cite{Reiter87}. The reasons for 
its widespread adoption are that
\begin{enumerate*}[label=\emph{(\roman*)}]
	\item \emph{it is broadly applicable}, because all it needs is a system description in some 
	monotonic knowledge representation language for which 
	a sound and complete inference method exists,
	\item \emph{it is sound and complete}, as it computes \emph{only} minimal diagnoses and can, in principle, output \emph{all} minimal diagnoses, and
	\item \emph{it computes diagnoses in best-first order} according to a given preference criterion.
\end{enumerate*}

However, HS-Tree per se does not encompass any specific provisions for being used in an iterative way. In other words, the DPI to be solved is assumed constant throughout the execution of HS-Tree. As a consequence of that, the question we address in this work is whether HS-Tree can be optimized for adoption \emph{in a sequential diagnosis scenario}, where the DPI to be solved is subject to successive change (information acquisition through measurements).
Already Raymond Reiter, in his seminal paper \cite{Reiter87} from 1987, 
asked: 
\begin{quoting}[font=itshape,leftmargin=10pt,vskip=3pt]
	When new diagnoses do arise as a result of system measurements, can we determine these new diagnoses in a reasonable way from the (\dots) HS-Tree already computed in determining the old diagnoses?
\end{quoting}
To the best of our knowledge, no algorithm has yet been proposed that sheds light on this very question.

As a result, state-of-the-art sequential approaches which draw on HS-Tree for diagnosis computation
handle the varying DPI to be solved by re-invoking HS-Tree each time a new 
piece of system knowledge (measurement outcome)
is obtained. This amounts to a \emph{discard-and-rebuild} usage of HS-Tree, where the search tree produced in one iteration is dropped prior to the next iteration, where a new 
one
is built from scratch. 
As the new tree obtained after incorporating the information about one measurement outcome usually quite closely resembles the existing tree, this approach generally requires substantial redundant computations, which often involve a significant number of expensive reasoner calls.
For instance, when debugging 
knowledge bases written in highly expressive logics such as OWL~2~DL \cite{grau2008owl}, a single 
call to 
an inference service is already 2NEXPTIME-complete.

Motivated by that, we propose DynamicHS, a \emph{stateful} variant of HS-Tree 
that pursues a \emph{reuse-and-adapt} strategy and is able to manage the dynamicity of the DPI throughout sequential diagnosis while avoiding the mentioned redundancy issues. The idea is to maintain \emph{one} (successively refined) tree data structure and to exploit the
information it contains to enhance computations in the subsequent iteration(s), e.g., by circumventing costly reasoner invocations. 
%
The main objective of DynamicHS is to allow for a better (time-)efficiency than HS-Tree while maintaining all aforementioned advantages (generality, soundness, completeness, best-first property) of the latter. Preliminary results of ongoing evaluations on several KB debugging problems---a domain where HS-Tree is the prevalent means for diagnosis computation \cite{friedrich2005gdm,Shchekotykhin2012,Kalyanpur2006a,meilicke2011,Horridge2011a}---suggest a superiority of DynamicHS against HS-Tree.

\section{Preliminaries}
\label{sec:basics}
We briefly characterize technical concepts used throughout this work, based on the framework of \cite{Shchekotykhin2012,Rodler2015phd} which is (slightly) more general \cite{rodler17dx_reducing} than Reiter's theory of diagnosis \cite{Reiter87}. 

\noindent\textbf{Diagnosis Problem Instance (DPI).} We assume that the diagnosed system, consisting of a set of components $\setof{c_1,\dots,c_k}$, is described by a finite set of logical sentences $\mo \cup \mb$, where $\mo$ (possibly faulty sentences) includes knowledge about the behavior of the system components, and $\mb$ (correct background knowledge) comprises any additional available 
system knowledge and system observations. More precisely, there is a one-to-one relationship between sentences $\tax_i \in \mo$ and components $c_i$, where $\tax_i$ describes the nominal behavior
of $c_i$ (\emph{weak fault model}). E.g., if $c_i$ is an AND-gate in a circuit, then $\tax_i := out(c_i) = and(in1(c_i),in2(c_i))$; $\mb$ in this example might encompass sentences stating, e.g., which components are connected by wires, 
or observed outputs of the circuit. The inclusion of a sentence $\tax_i$ in $\mo$ corresponds to the (implicit) assumption that $c_i$ is healthy. Evidence about the system behavior is captured by sets of positive ($\Tp$) and negative ($\Tn$) measurements \cite{Reiter87,dekleer1987,DBLP:journals/ai/FelfernigFJS04}. Each measurement is a logical sentence; positive ones $\tp\in\Tp$ must be true and negative ones $\tn\in\Tn$ must not be true. The former can be, depending on the context, e.g., observations about the system, probes or required system properties. The latter model, e.g., properties that must not hold for the system. We call $\tuple{\mo,\mb,\Tp,\Tn}$ a \emph{diagnosis problem instance (DPI)}. 

\begin{table}[t]
	\centering
	\scriptsize
	\renewcommand{\arraystretch}{1}
	\begin{tabular}{@{}ccc@{}}
		\toprule
		\multicolumn{1}{ c  }{\multirow{2}{*}{$\mo\;=$} } & \multicolumn{2}{ l  }{$\{ \tax_1: A \to \lnot B$ \;\; $\tax_2: A \to B$ \;\; $\tax_3: A \to \lnot C$} \\
		& \multicolumn{2}{ l  }{$\phantom{\{} \tax_4: B \to C$ \;\;\,\hspace{4pt} $\tax_5: A \to B \lor C \qquad\qquad\,\quad\;\;\;\}$} \\
		\cmidrule{1-3}
		\multicolumn{3}{ l  }{$\mb =\emptyset \quad\qquad\qquad\qquad\qquad \Tp=\emptyset \quad\qquad\qquad\qquad\qquad \Tn=\setof{\lnot A}$} \\
		
		\bottomrule
	\end{tabular}
	\caption{\small Example DPI stated in propositional logic.}
	\label{tab:example_DPI}
\end{table}

\begin{example}\label{ex:dpi}
Tab.~\ref{tab:example_DPI} depicts an example of a DPI, formulated in propositional logic. 
The ``system'' (which is the knowledge base itself in this case) comprises five ``components'' $c_1, \dots,c_5$, and the ``nominal behavior'' of $c_i$ is given by the respective axiom $\tax_i \in \mo$. There is neither any background knowledge ($\mb = \emptyset$) nor any positive test cases ($\Tp=\emptyset$) available from the start. But, there is one negative test case (i.e., $\Tn = \setof{\lnot A}$), which postulates that $\lnot A$ must \emph{not} be an entailment of the correct system (knowledge base). Note, however, that $\mo$ (i.e., the assumption that all ``components'' work nominally) in this case does entail $\lnot A$ (e.g., due to the axioms $\tax_1,\tax_2$) and therefore some axiom in $\mo$ must be faulty (i.e., some ``component'' is not healthy).\qed
\end{example}

\noindent\textbf{Diagnoses.} Given that the system description along with the positive measurements (under the 
assumption $\mo$ that all components are healthy) is inconsistent, i.e., $\mo \cup \mb \cup \Tp \models \bot$, or some negative measurement is entailed, i.e., $\mo \cup \mb \cup \Tp \models \tn$ for some $\tn \in \Tn$, some assumption(s) about the healthiness of components, i.e., some sentences in $\mo$, must be retracted. We call such a set of sentences $\md \subseteq \mo$ a \emph{diagnosis} for the DPI $\tuple{\mo,\mb,\Tp,\Tn}$ iff $(\mo \setminus \md) \cup \mb \cup \Tp \not\models x$ for all $x \in \Tn \cup \setof{\bot}$. We say that $\md$ is a \emph{minimal diagnosis} for $\dpi$ iff there is no diagnosis $\md' \subset \md$ for $\dpi$. The set of minimal diagnoses is representative for all diagnoses (under the weak fault model \cite{Kleer1992}), i.e., 
the set of all diagnoses is exactly given by the set of all supersets of all minimal diagnoses.
Therefore, diagnosis approaches usually restrict their focus to only minimal diagnoses. In the following, we denote the set of all minimal diagnoses for a DPI $\dpi$ by $\sol(\dpi)$. We furthermore denote by $\dt$ the \emph{actual diagnosis} which pinpoints the actually faulty axioms, i.e., all elements of $\dt$ are in fact faulty and all elements of $\mo\setminus\dt$ are in fact correct.

\begin{example}\label{ex:diagnoses}
For our DPI $\dpi$ in Tab.~\ref{tab:example_DPI} we have four minimal diagnoses, given by $\md_1:=[\tax_1,\tax_3]$, $\md_2:=[\tax_1,\tax_4]$, $\md_3:=[\tax_2,\tax_3]$, and $\md_4 := [\tax_2,\tax_5]$, i.e., $\sol(\dpi) = \setof{\md_1,\dots,\md_4}$.\footnote{In this work, we always denote diagnoses by squared brackets.} For instance, $\md_1$ is a minimal diagnosis as $(\mo\setminus\md_1) \cup \mb\cup \Tp = \setof{\tax_2,\tax_4,\tax_5}$ is both consistent and does not entail the given negative test case $\lnot A$.\qed
\end{example}

\noindent\textbf{Conflicts.} 
Instrumental for the computation of diagnoses is the concept of a conflict \cite{dekleer1987,Reiter87}.
A conflict is a set of healthiness assumptions for components $c_i$ that cannot all hold given the current knowledge about the system. More formally, $\mc \subseteq \mo$ is a \emph{conflict} for the DPI $\tuple{\mo,\mb,\Tp,\Tn}$ iff $\mc \cup \mb \cup \Tp \models x$ for some $x \in \Tn \cup \setof{\bot}$. We call $\mc$ a \emph{minimal conflict} for $\dpi$ iff there is no conflict $\mc' \subset \mc$ for $\dpi$.
In the following, we denote the set of all minimal conflicts for a DPI $\dpi$ by $\conf(\dpi)$.
A (minimal) diagnosis for $\dpi$ is then a (minimal) hitting set of all minimal conflicts for $\dpi$ \cite{Reiter87}, where $X$ is a \emph{hitting set} of a collection of sets $\mathbf{S}$ iff $X \subseteq \bigcup_{S_i \in \mathbf{S}} S_i$ and $X \cap S_i \neq \emptyset$ for all $S_i \in S$.

\begin{example}\label{ex:conflicts}
For our running example, $\dpi$, in Tab.~\ref{tab:example_DPI}, there are four minimal conflicts, given by $\mc_1 := \langle\tax_1,\tax_2\rangle$, $\mc_2 := \langle\tax_2,\tax_3,\tax_4\rangle$, $\mc_3 := \langle\tax_1,\tax_3,\tax_5\rangle$, and $\mc_4 := \langle\tax_3,\tax_4,\tax_5\rangle$, i.e., $\conf(\dpi) = \setof{\mc_1,\dots,\mc_4}$.\footnote{In this work, we always denote conflicts by angle brackets.} For instance, $\mc_4$, in CNF equal to $(\lnot A \lor \lnot C) \land (\lnot B \lor C) \land (\lnot A \lor B \lor C)$, is a conflict because, adding the unit clause $(A)$ to this CNF yields a contradiction, which is why the negative test case $\lnot A$ is an entailment of $\mc_4$. The minimality of the conflict $\mc_4$ can be verified by rotationally removing from $\mc_4$ a single axiom at the time and controlling for each so obtained subset that this subset is consistent and does not entail $\lnot A$.

For example, the minimal diagnosis $\md_1$ (see Example~\ref{ex:diagnoses}) is a hitting set of all minimal conflict sets because each conflict in $\conf(\dpi)$ contains $\tax_1$ or $\tax_3$. It is moreover a \emph{minimal} hitting set since the elimination of $\tax_1$ implies an empty intersection with, e.g., $\mc_1$, and the elimination of $\tax_3$ means that, e.g., $\mc_4$ is no longer hit.\qed   
\end{example}

\noindent\textbf{Sequential Diagnosis.}
Given multiple minimal diagnoses for a DPI, a sequential diagnosis process can be initiated. It 
involves a recurring execution of four phases,
\begin{enumerate*}[label=\emph{(\roman{*})}]
	\item computation of a set of leading (e.g., most probable) minimal diagnoses,
	\item selection of the best
	measurement based on these,
	\item conduction of measurement actions, and
	\item exploitation of the measurement outcome to refine the system knowledge.
\end{enumerate*} 
The goal in sequential diagnosis is to achieve sufficient diagnostic certainty (e.g., a single or highly probable remaining diagnosis) with highest efficiency. 
At this, the overall efficiency 
is determined by the costs 
for \emph{measurement conduction} and 
for \emph{computations of the diagnosis engine}. 
Whereas the former---which is not the topic of this work---can be ruled by proposing appropriate (low-cost, informative) measurements 
\cite{dekleer1987,pattipati1990,Shchekotykhin2012,Rodler2013,DBLP:journals/corr/Rodler2017,DBLP:conf/ruleml/RodlerS18,DBLP:conf/ieaaie/RodlerE19}, the latter is composed of the time required for 
\emph{diagnosis computation}, for 
\emph{measurement 
	selection},
as well as for the 
\emph{system knowledge update}.
We address the efficiency optimization problem in sequential diagnosis by suggesting new methods (DynamicHS algorithm) for the diagnosis computation and knowledge update processes. 
\begin{example}\label{ex:seq_diag}
Let $\tuple{\mo,\mb,\Tp,\Tn}$ be the example DPI described in Tab.~\ref{tab:example_DPI}, and assume that $\dt = \md_2$, i.e., $[\tax_1,\tax_4]$ is the actual diagnosis. Then, given the goal to obtain full diagnostic certainty (a single possible remaining diagnosis), one solution to the sequential diagnosis problem 
is given by 
$\Tp'=\{A \land B\}$ along with 
$\Tn'=\{C\}$. 
That is,
after adding these two measurements to the DPI, there is only a single minimal diagnosis ($\md_2$) left, whereas all others have been invalidated; formally: $\sol(\tuple{\mo,\mb,\Tp\cup\Tp',\Tn\cup\Tn'}) = \{\md_2\}$. 
\qed
\end{example}

\begin{algorithm}[t]
	\scriptsize
	\caption{Sequential Diagnosis}\label{algo:sequential_diagnosis}
	\begin{algorithmic}[1]
		\Require DPI $\dpi_0 := \langle\mo,\mb,\Tp,\Tn\rangle$, 
		probability measure $\pr$, 
		number $\ld$ ($\geq 2$) of minimal diagnoses to be computed per iteration, 
		heuristic $\qqm$ for measurement selection, boolean $\mathit{dynamic}$ 
		(use DynamicHS if true, HS-Tree otherwise)  
		\Ensure $\setof{\md}$ where $\md$ is the only remaining diagnosis for the extended DPI $\langle\mo,\mb,\Tp\cup\Tp',\Tn\cup\Tn'\rangle$ 
		\vspace{6pt}
		
		\State $\Tp'\gets \emptyset, \Tn'\gets \emptyset$ \label{algoline:inter_onto_debug:var_init_start} \Comment{performed measurements}
		\State $\mD_{\checkmark} \gets \emptyset, \mD_{\times} \gets \emptyset$,  
		$\state \gets \tuple{[\,[]\,],[\,],\emptyset,\emptyset}$ \label{algoline:inter_onto_debug:var_init_end} \Comment{initial state of DynamicHS}
		\While{\true}  \label{algoline:inter_onto_debug:while}
		\If{$\mathit{dynamic}$}
			\State $\tuple{\mD,\state} \gets \Call{DynamicHS}{\dpi_0,\Tp',\Tn',\pr, \ld, \mD_{\checkmark}, \mD_{\times}, \state}$
			\label{algoline:inter_onto_debug:call_DynHS}
		\Else
			\State $\mD \gets \Call{HS-Tree}{\dpi_0,\Tp',\Tn',\pr, \ld}$
			\label{algoline:inter_onto_debug:call_HS-Tree}
		\EndIf
		\IfThen {$|\mD|=1$}
		{\Return $\mD$}
		\label{algoline:inter_onto_debug:if_goal_reached}
		\State $mp \gets \Call{computeBestMeasPoint}{\mD,\dpi_0,\Tp',\Tn', \pr, \qqm}$   \label{algoline:inter_onto_debug:calc_query} 
		\State $\mathit{outcome} \gets \Call{performMeas}{mp}$   \Comment{oracle inquiry (user interaction)} \label{algoline:inter_onto_debug:perform_meas}
		\State $\tuple{\Tp', \Tn'} \gets \Call{addMeas}{mp,outcome,\Tp', \Tn'}$ 
		\label{algoline:inter_onto_debug:add_meas}
		\If{$\mathit{dynamic}$}
			\State $\tuple{\mD_{\checkmark}, \mD_{\times}} \gets \Call{assignDiagsOkNok}{\mD,\dpi_0,\Tp', \Tn'}$
			\label{algoline:inter_onto_debug:partition_D_into_Dcheckmark_and_Dtimes}
		\EndIf
		\EndWhile
	\end{algorithmic}
	\normalsize
\end{algorithm}


\begin{algorithm}[h!]
	\scriptsize
	\caption{DynamicHS} \label{algo:dynamic_hs}
	\begin{algorithmic}[1]
		\Require 
		\textcolor{white}{.}
		tuple $\tuple{\dpi, \Tp', \Tn',  \pr, \ld, \mD_{\checkmark}, \mD_{\times},  \state}$ comprising
		\begin{itemize}[noitemsep]
			\item a DPI $\dpi = \langle\mo,\mb,\Tp,\Tn\rangle$
			\item the acquired sets of positive ($\Tp'$) and negative ($\Tn'$) measurements so far 
			\item a function $\pr$ assigning a fault probability to each element in $\mo$
			\item the number $\ld$ of leading minimal diagnoses to be computed 
			\item the set $\mD_{\checkmark}$ of all elements of the set $\mD_{calc}$ (returned by the previous \textsc{dynamicHS} run) which are minimal diagnoses wrt.\ $\langle\mo,\mb,\Tp\cup\Tp',\Tn\cup\Tn'\rangle$
			\item the set $\mD_{\times}$ of all elements of the set $\mD_{calc}$ (returned by the previous \textsc{dynamicHS} run) which are no diagnoses wrt.\ $\langle\mo,\mb,\Tp\cup\Tp',\Tn\cup\Tn'\rangle$
			\item $\state = \tuple{\Queue, \Queue_{dup}, \mD_{\supset}, \mC_{\mathit{calc}}}$ where
			\begin{itemize}[noitemsep]
				\item $\Queue$ is the current queue of unlabeled nodes,
				\item $\Queue_{dup}$ is the current queue of duplicate nodes,
\item $\mD_{\supset}$ is the current set of computed non-minimal diagnoses,
\item $\mC_{calc}$ is the current set of computed minimal conflict sets.
			\end{itemize}
		\end{itemize}
		\Ensure 
		tuple $\tuple{\mD,\state}$ where
		\begin{itemize}[noitemsep]
			\item $\mD$ is the set of the $\ld$ (if existent) most probable (as per $\pr$) minimal diagnoses wrt.\
			$\langle\mo,\mb,\Tp\cup\Tp',\Tn\cup\Tn'\rangle$
			\item $\state$ is as described above
		\end{itemize}  
		
		\vspace{6pt}
		\Procedure{dynamicHS}{$\dpi, \Tp', \Tn',  \pr, \ld, \mD_{\checkmark}, \mD_{\times}, \state $}
		\State $\mD_{calc} \gets \emptyset$\label{algoline:dyn:Dcalc_gets_emptyset}
		\State $\state \gets \Call{updateTree}{\dpi, \Tp', \Tn',  \pr, \mD_{\checkmark}, \mD_{\times}, \state}$\label{algoline:dyn:update_tree}     
		\While{$\Queue \neq [\,] \land \left(\;|\mD_{calc}| < \ld\;\right)$}	
		\label{algoline:dyn:while}																																
		\State $\mathsf{node} \gets \Call{getAndDeleteFirst}{\Queue}$\label{algoline:dyn:get_first}			  \Comment{$\mathsf{node}$ is processed}		
		\If{$\mathsf{node} \in \mD_{\checkmark}$}\label{algoline:dyn:node_in_Dcheckmark}  \Comment{$\mD_{\checkmark}$ includes only min...}
		\State $L \gets \mathit{valid}$\label{algoline:dyn:set_L_to_valid}      \Comment{...diags wrt.\ current DPI}         
		\Else
		\State $\tuple{L,\state} \gets \Call{dLabel}{\dpi, \Tp', \Tn', \pr, \mathsf{node}, \mD_{calc}, \state}$\label{algoline:dyn:dlabel}
		\EndIf
		\If{$L = \mathit{valid}$}\label{algoline:dyn:if_L_valid}  
		\State $\mD_{calc} \gets \mD_{calc} \cup \setof{\mathsf{node}}$\label{algoline:dyn:add_to_Dcalc}      \Comment{$\mathsf{node}$ is a min diag wrt.\ current DPI}
		\ElsIf{$L = \mathit{nonmin}$}								
		\State $\mD_{\supset} \gets \mD_{\supset} \cup \setof{\mathsf{node}}$ \label{algoline:dyn:add_to_Dsupset}  \Comment{$\mathsf{node}$ is a non-min diag wrt.\ current DPI}
		\Else 	
		\For{$e \in L$}\label{algoline:dyn:for_e_in_L}            \Comment{$L$ is a min conflict  wrt.\ current DPI}
		\State $\mathsf{node}_e \gets \Call{append}{\mathsf{node},e}$ \label{algoline:dyn:add_ax_to_node}       \Comment{$\mathsf{node}_e$ is generated}   
		\State $\mathsf{node}_{e}.\mathsf{cs} \gets \Call{append}{\mathsf{node.cs},L}$ \label{algoline:dyn:add_cs_to_node.cs}
		\If{$\mathsf{node}_e \in \Queue$}   \label{algoline:dyn:check_node_already_in_Q}                      \Comment{$\mathsf{node}_e$ is (\emph{set-equal}) duplicate of node in $\Queue$}
		\State $\Queue_{dup} \gets \Call{insertSorted}{ \mathsf{node}_e, \Queue_{dup}, \mathit{card}, <}$ 
		\label{algoline:dyn:add_to_Qdup}
		\Else
		\State $\Queue \gets \Call{insertSorted}{ \mathsf{node}_e, \Queue, \pr, >}$\label{algoline:dyn:add_to_Queue} 
		\EndIf
		\EndFor
		\EndIf
		\EndWhile
		\State \Return $\tuple{\mD_{calc}, \state}$ \label{algoline:dyn:return}
		\EndProcedure
		\vspace{6pt}
		\Procedure{\textsc{dLabel}}{$\langle\mo,\mb,\Tp,\Tn\rangle,\Tp', \Tn', \pr, \mathsf{node}, \mD_{calc}, \state$} 
		\For{$\mathsf{nd} \in \mD_{calc}$}\label{algoline:dlabel:non-min_crit_start}
		\If{$\mathsf{node} \supset \mathsf{nd}$}    \Comment{$\mathsf{node}$ is a non-min diag}
		\State \Return $\tuple{\mathit{nonmin},\state}$
		\EndIf
		\EndFor\label{algoline:dlabel:non-min_crit_end}
		\For{$\mc \in \mathbf{C}_{calc}$}\label{algoline:dlabel:reuse_start} \Comment{$\mC_{calc}$ includes conflicts wrt.\ current DPI}
		\If{$\mc \cap \mathsf{node} = \emptyset$}\label{algoline:dlabel:if_C_cap_node=emptyset}    \Comment{reuse (a subset of) $\mc$ to label $\mathsf{node}$} 
		\State $X \gets \Call{findMinConflict}{\langle\mc,\mb,\Tp\cup\Tp',\Tn\cup\Tn'\rangle}$\label{algoline:dlabel:qx_1} 
		\If{$X = \mc$} \label{algoline:dlabel:if_X=C}
		\State \Return $\tuple{\mc,\state}$\label{algoline:dlabel:return_C}
		\Else      \Comment{$X \subset \mc$} \label{algoline:dlabel:else}
		\State $\state \gets \Call{prune}{X,\state,\pr}$ \label{algoline:dlabel:prune}
		\State \Return $\tuple{X,\state}$ \label{algoline:dlabel:return_X}
		\EndIf
		\EndIf
		\EndFor\label{algoline:dlabel:reuse_end}
		\State $L\gets \Call{findMinConflict}{\langle\mo\setminus\mathsf{node},\mb,\Tp\cup\Tp',\Tn\cup\Tn'\rangle}$\label{algoline:dlabel:qx_2} 
		\If{$L$ = \text{'no conflict'}}						\Comment{$\mathsf{node}$ is a diag}
		\State \Return $\tuple{\mathit{valid},\state}$\label{algoline:dlabel:return_valid}
		\Else						\Comment{$L$ is a \emph{new} min conflict ($\notin \mathbf{C}_{calc}$)}
		\State $\mathbf{C}_{calc} \gets \mathbf{C}_{calc} \cup \setof{L}$\label{algoline:dlabel:add_new_cs}
		\State \Return $\tuple{L,\state}$\label{algoline:dlabel:return_new_cs}
		\EndIf
		\EndProcedure
%
\vspace{6pt}
\Procedure{\textsc{updateTree}}{$\dpi, \Tp', \Tn',  \pr, \mD_{\checkmark}, \mD_{\times}, \state$}
\For{$\mathsf{nd} \in \mD_{\times}$} \label{algoline:update:process_Dtimes_start}
\Comment{search for redundant nodes among invalidated diags}
\If{$\Call{redundant}{\mathsf{nd},\dpi}$}  \label{algoline:update:call_redundant_function}
	\State $\state \gets \Call{prune}{X,\state,\pr}$ \label{algoline:update:prune}
\EndIf
\EndFor
\For{$\mathsf{nd} \in \mD_{\times}$}\label{algoline:update:reinsert_D_of_Dx_to_Q}
\Comment{add all (non-pruned) nodes in $\mD_{\times}$ to $\Queue$}
\State $\Queue \gets \Call{insertSorted}{ \mathsf{nd}, \Queue, \pr, >}$\label{algoline:update:insert_sorted_0}
\State $\mD_{\times} \gets \mD_{\times} \setminus \setof{\mathsf{nd}}$ \label{algoline:update:delete_from_Dtimes}
\EndFor \label{algoline:update:process_Dtimes_end}
\For{$\mathsf{nd} \in \mD_{\supset}$}\label{algoline:update:process_Dsupset_start} \Comment{add all (non-pruned) nodes in $\mD_{\supset}$ to $\Queue$, which...}
\State $\mathit{nonmin} \gets \false$ \Comment{...are no longer supersets of any diag in $\mD_{\checkmark}$}
\For{$\mathsf{nd}' \in \mD_{\checkmark}$}
\If{$\mathsf{nd} \supset \mathsf{nd}'$}    
\State $\mathit{nonmin} \gets \true$
\State \textbf{break} 
\EndIf
\EndFor
\If{$\mathit{nonmin} = \false$}
\State $\Queue \gets \Call{insertSorted}{ \mathsf{nd}, \Queue, \pr, >}$\label{algoline:update:insert_sorted_0.5}
\State $\mD_{\supset} \gets \mD_{\supset} \setminus \setof{\mathsf{nd}}$\label{algoline:update:delete_from_Dsupset}
\EndIf
\EndFor \label{algoline:update:process_Dsupset_end}
\For{$\md \in \mD_{\checkmark}$}\label{algoline:update:process_Dcheckmark_start}  \Comment{add known min diags in $\mD_{\checkmark}$ to $\Queue$ to find diags...}
\State $\Queue \gets \Call{insertSorted}{ \md, \Queue, \pr, >}$\label{algoline:update:insert_sorted_1}
\Comment{...in best-first order (as per $\pr$)}
\EndFor \label{algoline:update:process_Dcheckmark_end}
\State \Return $\state$
\EndProcedure	
\end{algorithmic}
	\normalsize
\end{algorithm}



\section{DynamicHS Algorithm}

\noindent\textbf{Inputs and Outputs.} 
DynamicHS (Alg.~\ref{algo:dynamic_hs}) accepts the following arguments: 
(1)~an initial DPI $\dpi_0 = \tuple{\mo,\mb,\Tp,\Tn}$,
(2)~the already accumulated positive and negative measurements $\Tp'$ and $\Tn'$, 
(3)~a probability measure $\pr$ (allowing to compute the probability of diagnoses), 
(4)~a stipulated number $\ld\geq 2$ of diagnoses to be returned, 
(5)~the set of those diagnoses returned by the previous DynamicHS run that are consistent ($\mD_{\checkmark}$) and those that are inconsistent ($\mD_{\times}$) with the latest added measurement, 
and
(6)~a tuple of variables $\state$ (cf.\ Alg.~\ref{algo:dynamic_hs}), which altogether describe DynamicHS's current state.
It outputs the $\ld$ (if existent) minimal diagnoses of maximal probability wrt.\ $\pr$ for the DPI 
$\langle\mo,\mb,\Tp \cup \Tp',\Tn \cup \Tn'\rangle$.

\noindent\textbf{Embedding in Sequential Diagnosis Process.} Alg.~\ref{algo:sequential_diagnosis} sketches a generic sequential diagnosis algorithm and shows how it accommodates DynamicHS (line~\ref{algoline:inter_onto_debug:call_DynHS}) or, alternatively, Reiter's HS-Tree (line~\ref{algoline:inter_onto_debug:call_HS-Tree}), as methods for iterative diagnosis computation. 
The algorithm 
reiterates a while-loop (line~\ref{algoline:inter_onto_debug:while}) 
until the solution space of minimal diagnoses includes only a single element.\footnote{
	Of course, less rigorous stopping criteria are possible, e.g., when a diagnosis exceeds a predefined probability threshold \cite{dekleer1987}.
} 
This is the case iff a diagnoses set $\mD$ with $|\mD| = 1$ is output (line~\ref{algoline:inter_onto_debug:if_goal_reached}) since both DynamicHS and HS-Tree are complete 
and always attempt to compute 
at least two
diagnoses ($\ld \geq 2$). On the other hand, as long as $|\mD| > 1$, the algorithm 
seeks to acquire additional information
to rule out further elements in $\mD$.
To this end, the best next measurement point $mp$ is computed (line~\ref{algoline:inter_onto_debug:calc_query}), using the current system information---$\dpi_0$, $\mD$, and acquired measurements $\Tp'$, $\Tn'$---as well as the given probabilistic information $\pr$ and some measurement selection heuristic $\qqm$ (which defines what ``best'' means, cf. \cite{rodler17dx_activelearning}).
%
The conduction of the measurement at $mp$ (line~\ref{algoline:inter_onto_debug:perform_meas}) is usually accomplished by a qualified 
user 
(\emph{oracle}) that interacts with the sequential diagnosis system, e.g., an electrical engineer for a defective circuit, or a domain expert in case of a faulty ontology.
The measurement point $mp$ along with its result $\mathit{outcome}$ are used to formulate a logical sentence $m$ that is either added to $\Tp'$ if $m$ constitutes a positive measurement, and to $\Tn'$ otherwise (line~\ref{algoline:inter_onto_debug:add_meas}).
Finally, if DynamicHS is adopted, 
the set of diagnoses $\mD$ is partitioned 
into those consistent ($\mD_{\checkmark}$) and those inconsistent ($\mD_{\times}$) with the newly added measurement $m$ (line~\ref{algoline:inter_onto_debug:partition_D_into_Dcheckmark_and_Dtimes}).

\noindent\textbf{Reiter's HS-Tree.} DynamicHS inherits many of its aspects from Reiter's HS-Tree.
Hence, we first recapitulate HS-Tree and then focus on the differences to and idiosyncrasies of DynamicHS.

HS-Tree computes minimal diagnoses for $\dpi=\langle\mo,\mb$, $\Tp\cup\Tp',\Tn\cup\Tn'\rangle$ in a sound, complete\footnote{Unlike Reiter, we assume that only \emph{minimal} conflicts are used as node labels. Thus, the issue pointed out by \cite{greiner1989correction} does not arise.} and best-first way.
%
Starting from a priority queue of unlabeled nodes $\Queue$, initially
comprising only an unlabeled root node, 
the algorithm continues to remove and label the first ranked node from $\Queue$ (\textsc{getAndDeleteFirst})
until all nodes are labeled ($\Queue = [\,]$) or $\ld$ minimal diagnoses have been computed.
The possible node labels are minimal conflicts (for internal tree nodes) and 
$\mathit{valid}$
as well as 
$\mathit{closed}$
(for leaf nodes). 
All minimal conflicts that have already been computed and used as node labels are stored in the (initially empty) set $\mC_{calc}$. 
Each edge in the constructed tree has a label. For ease of notation, each tree node $\mathsf{nd}$ is conceived of as the set of edge labels along the branch from the root node to itself. 
%
E.g., the node at location $\textcircled{\tiny 12}$ in iteration~1 of Fig.~\ref{fig:hs_example} is referred to as $\setof{1,2,5}$.
Once the tree has been completed ($\Queue = [\,]$), i.e., there are no more unlabeled nodes, it holds that 
$\sol(\dpi) = \setof{\nd\mid \nd \text{ is labeled by }\mathit{valid}}$.

To label a node $\nd$, the algorithm calls a labeling function (\textsc{label}) 
which executes the following tests in the given order and returns as soon as a label for $\nd$ has been determined:
\begin{enumerate}[noitemsep,leftmargin=*,label=(L\arabic*)]
	\item \label{enum:hstree:label:non-min} \emph{(non-minimality):} Check if $\nd$ is non-minimal (i.e.\ whether there is a node $\node$ with label 
	$\mathit{valid}$
	where $\nd \supseteq \node$). If so, $\nd$ is labeled by 
	$\mathit{closed}$.
	\item \label{enum:hstree:label:duplicate} \emph{(duplicate):} Check if $\nd$ is duplicate (i.e.\ whether $\nd = \node$ for some other $\node$ in $\Queue$). If so, $\nd$ is labeled by 
	$\mathit{closed}$.
	\item \label{enum:hstree:label:conflict_reuse} \emph{(reuse label):} Scan $\mC_{calc}$ for some $\mc$ such that $\nd \cap \mc = \emptyset$. If so, $\nd$ is labeled by $\mc$. 
	\item \label{enum:hstree:label:getMinConflict} \emph{(compute label):} Invoke \textsc{getMinConflict}, a \emph{(sound and complete) minimal conflict searcher}, e.g., QuickXPlain \cite{junker04}, to get a minimal conflict for $\tuple{\mo\setminus\nd,\mb,\Tp\cup\Tp',\Tn\cup\Tn'}$. If a minimal conflict $\mc$ is output, $\nd$ is labeled by $\mc$.
	Otherwise, if 'no conflict' is returned, then $\nd$ is labeled by 
	$\mathit{valid}$.
\end{enumerate}
All nodes labeled by 
$\mathit{closed}$
or 
$\mathit{valid}$
have no successors and are leaf nodes.
For each node $\nd$ labeled by a minimal conflict $L$, 
one outgoing edge is constructed for each element $e\in L$, where this edge is labeled by $e$ and pointing to a newly created unlabeled node $\nd \cup \setof{e}$. 
Each new node is added to $\Queue$ such that $\Queue$'s sorting is preserved (\textsc{insertSorted}).
$\Queue$ might be, e.g., \emph{(i)}~a FIFO queue, entailing that HS-Tree computes diagnoses in minimum-cardinality-first order (\emph{breadth-first search}), or \emph{(ii)}~sorted in descending order by $\pr$, where most probable diagnoses are generated first (\emph{uniform-cost search}; for details see \cite[Sec.~4.6]{Rodler2015phd}). 
%

Finally, note the \emph{statelessness} of Reiter's HS-Tree, reflected by $\Queue$ initially including \emph{only an unlabeled root node}, and $\mC_{calc}$ being initially \emph{empty}. That is, a HS-Tree is built from scratch in each iteration, every time for different measurement sets $\Tp',\Tn'$.

\noindent\textbf{Dynamicity of DPI in Sequential Diagnosis.}
In the course of sequential diagnosis (Alg.~\ref{algo:sequential_diagnosis}), where additional system knowledge is gathered in terms of measurements, 
the DPI is subject to gradual change---it is dynamic. 
At this, each addition of a new (informative) measurement 
to the DPI 
also effectuates a transition of the sets of (minimal) diagnoses and (minimal) conflicts. Whereas this fact is of no concern to a stateless diagnosis computation strategy, it has to be carefully taken into account when engineering a stateful approach.

\noindent\textbf{Towards Stateful Hitting Set Computation.}
To understand the necessary design decisions to devise a sound and complete stateful hitting set search, 
we look at 
more specifics of the conflicts and diagnoses evolution in
sequential diagnosis:\footnote{
	See \cite[Sec. 12.4]{Rodler2015phd} for a more formal treatment and proofs.
}
\begin{property}\label{property:impact_of_DPI_transition}
Let $\dpi_j = \tuple{\mo,\mb,\Tp,\Tn}$ be a DPI and let $T$ be 
Reiter's HS-Tree for $\dpi_j$ (executed until) producing the 
diagnoses $\mD$ where $|\mD|\geq 2$. 
Further, 
let $\dpi_{j+1}$ be the DPI resulting from $\dpi_j$ through the addition of an informative\footnote{That is, 
adding $m$ to 
the (positive or negative) measurements of the DPI effectuates an invalidation of some diagnosis in $\mD$.} measurement $m$ to either $\Tp$ or $\Tn$. Then:
\begin{enumerate}[leftmargin=*,noitemsep]
	\item \label{property:impact_of_DPI_transition:enum:T_must_be_updated} $T$ is not a correct HS-Tree for $\dpi_{j+1}$, i.e., (at least) some node labeled by $\mathit{valid}$ in $T$ is incorrectly labeled. \\
	\emph{(That is, to reuse $T$ for $\dpi_{j+1}$, $T$ must be updated.)}
	\item \label{property:impact_of_DPI_transition:enum:diags_can_only_grow} Each $\md \in \sol(\dpi_{j+1})$ is either equal to or a superset of some $\md' \in \sol(\dpi_{j})$. \\ 
	\emph{(That is, minimal diagnoses can grow or remain unchanged, but cannot shrink. Consequently, to reuse $T$ for sound and complete minimal diagnosis computation for $\dpi_{j+1}$, existing nodes must never be reduced---either a node is left as is, deleted as a whole, or (prepared to be) extended.)}
	\item \label{property:impact_of_DPI_transition:enum:conflicts_can_only_shrink} For all $\mc \in \conf(\dpi_{j})$ there is a $\mc' \in \conf(\dpi_{j+1})$ such that $\mc' \subseteq \mc$. \\
	\emph{(That is, existing minimal conflicts can only shrink or remain unaffected, but cannot grow. Hence, priorly computed minimal conflicts (for an old DPI) might not be minimal for the current DPI. In other words, conflict node labels of $T$ can, but do not need to be, correct for $\dpi_{j+1}$.)}
	\item \label{property:impact_of_DPI_transition:enum:some_conflict_shrinks_or_new_conflict_arises} (a) There is some $\mc \in \conf(\dpi_{j})$ for which there is a $\mc' \in \conf(\dpi_{j+1})$ with $\mc' \subset \mc$, and/or \\
	(b) there is some $\mc_{\mathit{new}} \in \conf(\dpi_{j+1})$ where $\mc_{\mathit{new}} \not\subseteq \mc$ and $\mc_{\mathit{new}} \not\supseteq \mc$ for all $\mc \in \conf(\dpi_{j})$. \\
	\emph{(That is, (a) some minimal conflict is reduced in size, and/or (b) some entirely new minimal conflict (not in any subset-relationship with existing ones) arises. Some existing node in $T$ which represents a minimal diagnosis for $\dpi_{j}$ (a) can be deleted since it would not be present when using $\mc'$ as node label in $T$ wherever $\mc$ is used, or (b) must be extended to constitute a diagnosis for $\dpi_{j+1}$, since it does not hit 
	$\mc_{\mathit{new}}$.)}
\end{enumerate}
\end{property}

\noindent\textbf{Major Modifications to Reiter's HS-Tree.}
Based on Property~\ref{property:impact_of_DPI_transition}, the following principal amendments to Reiter's HS-Tree are necessary to make it a properly-working stateful diagnosis computation method:
\begin{enumerate}[label=\textbf{(Mod\arabic*)}, wide, labelwidth=!, labelindent=0pt,itemsep=2pt]
	\item \label{enum:mod:nonmin_diags_and_duplicates_stored} Non-minimal diagnoses (test \ref{enum:hstree:label:non-min} in HS-Tree) and duplicate nodes (test \ref{enum:hstree:label:duplicate}) are stored in collections $\mD_{\supset}$ and $\Queue_{dup}$, respectively, instead of being closed and discarded.\vspace{2pt}\\ 
%
%
	\emph{Justification:} Property~\ref{property:impact_of_DPI_transition}.\ref{property:impact_of_DPI_transition:enum:diags_can_only_grow} suggests to store also non-minimal diagnoses, as they might constitute (sub-branches of) minimal diagnoses in next iteration. 
	Further, Property~\ref{property:impact_of_DPI_transition}.\ref{property:impact_of_DPI_transition:enum:some_conflict_shrinks_or_new_conflict_arises}(a) suggests to record all duplicates for completeness of the diagnosis search. Because, some active node $\mathsf{nd}$ representing this duplicate in the current tree could become obsolete due to the shrinkage of some conflict, and the duplicate might be non-obsolete and eligible to turn active and replace $\mathsf{nd}$ in the tree.
	
	\item \label{enum:mod:nodes_are_lists} Each node $\mathsf{nd}$ is no longer identified with the \emph{set} of edge labels along its branch,
	but as an \emph{ordered list} of these edge labels. In addition, an ordered list of the conflicts used to label internal nodes along this branch 
	is stored in terms of $\mathsf{nd.cs}$.
	E.g., for node $\mathsf{nd}$ at location $\textcircled{\scriptsize 9}$ in iteration~1 of Fig.~\ref{fig:dynhs_example}, $\mathsf{nd} = [2,5]$ and $\mathsf{nd.cs} = [\tuple{1,2},\tuple{1,3,5}]$.
	\vspace{2pt}\\
%
%
	\emph{Justification:} (Property~\ref{property:impact_of_DPI_transition}.\ref{property:impact_of_DPI_transition:enum:some_conflict_shrinks_or_new_conflict_arises}) The reason for storing both the edge labels and the internal node labels as lists 
	lies in the replacement of obsolete tree branches by stored duplicates. In fact, any duplicate used to replace a node must correspond to the same \emph{set} of edge labels as the replaced node. However, in the branch of the obsolete node, some node-labeling conflict has been reduced to make the node redundant, whereas for a suitable duplicate replacing the node, no redundancy-causing changes to conflicts along its branch have occurred. By storing only sets of edge labels, we could not differentiate between the redundant and the non-redundant nodes. 
%
%
	\item \label{enum:mod:conflict_minimality_check} Before reusing a conflict $\mc$ 
	(labeling test \ref{enum:hstree:label:conflict_reuse} in HS-Tree), a minimality check for $\mc$ is performed. 
	If this leads to the identification of a conflict $X\subset\mc$ for the current DPI,
	$X$ is used to prune obsolete tree branches, to replace node-labeling conflicts that are supersets of $X$,
	and to update $\mC_{calc}$ in that $X$ is added and all of its supersets are deleted.\vspace{2pt}\\
%
%
	\emph{Justification:} (Property~\ref{property:impact_of_DPI_transition}.\ref{property:impact_of_DPI_transition:enum:conflicts_can_only_shrink}) Conflicts in $\mC_{calc}$ and those appearing as labels in the existing tree (elements of lists $\mathsf{nd.cs}$ for different nodes $\mathsf{nd}$) might not be minimal for the current DPI (as they might have been computed for a prior DPI). This minimality check helps both to prune the tree (reduction of number of nodes) and to make sure that extensions to the tree use only minimal conflicts wrt.\ the current DPI as node labels (avoidance of the introduction of unnecessary new edges).
	
	\item \label{enum:mod:tree_update} Execution of a tree update at start of each DynamicHS execution, where the tree produced for a previous
	DPI is adapted to a tree that allows to compute minimal diagnoses for the current DPI in a sound, complete and best-first way.\vspace{2pt}\\
%
%
	\emph{Justification:} 
	Property~\ref{property:impact_of_DPI_transition}.\ref{property:impact_of_DPI_transition:enum:T_must_be_updated}. 
	
	\item \label{enum:mod:storing_the_state} State of DynamicHS (in terms of the so-far produced tree) is stored over all its iterations executed throughout sequential diagnosis (Alg.~\ref{algo:sequential_diagnosis}) by means of the tuple $\mathsf{state}$.\vspace{2pt}\\
	%
	%
	\emph{Justification:} Statefulness of DynamicHS. 
\end{enumerate}

\noindent\textbf{DynamicHS: Algorithm Walkthrough.}
%
Like HS-Tree, DynamicHS (Alg.~\ref{algo:dynamic_hs}) is processing a priority queue $\Queue$ of nodes (while-loop). In each iteration, the top-ranked node $\mathsf{node}$ is removed from $\Queue$ to be labeled (\textsc{getAndDeleteFirst}).    
Before calling the labeling function (\textsc{dLabel}), however, the algorithm checks if $\mathsf{node}$ is among the known minimal diagnoses $\mD_{\checkmark}$ from the previous iteration 
(line~\ref{algoline:dyn:node_in_Dcheckmark}). If so, the node is directly labeled by $\mathit{valid}$ (line~\ref{algoline:dyn:set_L_to_valid}). Otherwise \textsc{dLabel} is invoked to compute a label for $\mathsf{node}$ (line~\ref{algoline:dyn:dlabel}).\vspace{3pt}

\textsc{dLabel}: First, the non-minimality check is performed (lines~\ref{algoline:dlabel:non-min_crit_start}--\ref{algoline:dlabel:non-min_crit_end}), just as in \ref{enum:hstree:label:non-min} in HS-Tree.
If negative, a conflict-reuse check is carried out (lines~\ref{algoline:dlabel:reuse_start}--\ref{algoline:dlabel:reuse_end}). Note, the duplicate check (\ref{enum:hstree:label:duplicate} in HS-Tree)
is obsolete since no duplicate nodes 
can ever be elements of $\Queue$ in DynamicHS (duplicates are identified and added to $\Queue_{dup}$ at node generation time, lines \ref{algoline:dyn:check_node_already_in_Q}--\ref{algoline:dyn:add_to_Qdup}). 
The conflict-reuse check starts equally as in HS-Tree. However, if a conflict $\mc$ for reuse is found in 
$\mC_{calc}$ (line~\ref{algoline:dlabel:if_C_cap_node=emptyset}), then the minimality of $\mc$ wrt.\ the \emph{current} DPI is checked using \textsc{findMinConflict} (line~\ref{algoline:dlabel:qx_1}).
If a conflict $X\subset\mc$ is detected (line~\ref{algoline:dlabel:else}), 
then $X$ is used to prune the current hitting set tree (line~\ref{algoline:dlabel:prune}; \textsc{prune} function, see below).
Finally, the found minimal conflict ($\mc$ or $X$, depending on minimality check) is used to label $\mathsf{node}$ (lines~\ref{algoline:dlabel:return_C}, \ref{algoline:dlabel:return_X}).
%
The case where there is no conflict for reuse
is handled just as in HS-Tree (lines~\ref{algoline:dlabel:qx_2}--\ref{algoline:dlabel:return_new_cs}, cf.\ \ref{enum:hstree:label:getMinConflict}).
Finally, note that \textsc{dLabel} gets and returns the tuple $\mathsf{state}$ (current tree state) as an argument, since the potentially performed pruning actions (line~\ref{algoline:dlabel:prune})
might modify $\mathsf{state}$.\vspace{3pt}  

The output of \textsc{dLabel} is then processed by DynamicHS (lines~\ref{algoline:dyn:if_L_valid}--\ref{algoline:dyn:return})
Specifically,
$\mathsf{node}$ is assigned to $\mD_{calc}$ if the returned label is $\mathit{valid}$ (line~\ref{algoline:dyn:add_to_Dcalc}), and to $\mD_{\supset}$ if the label is $\mathit{nonmin}$ (line~\ref{algoline:dyn:add_to_Dsupset}). If
the label is a minimal conflict $L$, then a child node $\mathsf{node}_e$ is created for each element $e \in L$ and assigned to either $\Queue_{dup}$ (line~\ref{algoline:dyn:add_to_Qdup}) if there is a node in $\Queue$ that is \emph{set}-equal to $\mathsf{node}_e$, or to $\Queue$ otherwise (line~\ref{algoline:dyn:add_to_Queue}). At this, $\mathsf{node}_e$ is constructed from $\mathsf{node}$ via the \textsc{append} function (lines \ref{algoline:dyn:add_ax_to_node} and \ref{algoline:dyn:add_cs_to_node.cs}), which appends the element $e$ to the list $\mathsf{node}$, and the conflict $L$ to the list $\mathsf{node.cs}$ (cf.\ \ref{enum:mod:nodes_are_lists} above). 

When the hitting set tree 
has been completed ($\Queue = [\,]$), or $\ld$ diagnoses have been found ($|\mD_{calc}| = \ld$), DynamicHS returns $\mD_{calc}$ along with the current tree state $\mathsf{state}$.\vspace{3pt}
%

\textsc{updateTree}: The goal is to adapt the existing tree in a way it constitutes a basis for finding all and only minimal diagnoses in highest-probability-first order for the \emph{current} DPI. 
The strategy 
is to search for non-minimal conflicts to be updated, and tree branches to be pruned, among the
minimal diagnoses for the previous DPI
that have been invalidated by the latest added measurement (the elements of $\mD_{\times}$).

Regarding the pruning of tree branches, we call a node $\mathsf{nd}$ \emph{redundant} (wrt.\ a DPI $\dpi$) iff there is some index $j$ and a minimal conflict $X$ wrt.\ $\dpi$ such that the conflict $\mathsf{nd.cs}[j] \supset X$ and the element $\mathsf{nd}[j] \in \mathsf{nd.cs}[j] \setminus X$. Moreover, we call $X$ a \emph{witness of redundancy for $\mathsf{nd}$} (wrt.\ $\dpi$).
Simply put, $\mathsf{nd}$ is redundant iff 
it would not exist given that the (current) minimal conflict $X$ had been used as a label instead of the (old, formerly minimal, but by now) non-minimal conflict $\mathsf{nd.cs}[j]$.

If a redundant node is detected among the elements of $\mD_{\times}$ (function \textsc{redundant}),
then the \textsc{prune} function (see below) is called given the witness of redundancy of the redundant node as an argument (lines \ref{algoline:update:process_Dtimes_start}--\ref{algoline:update:prune}). After each node in $\mD_{\times}$ has been processed,
the remaining nodes in $\mD_{\times}$ (those that are non-redundant and thus have not been pruned) are re-added to $\Queue$ in prioritized order (\textsc{insertSorted}) according to $\pr$ (lines \ref{algoline:update:reinsert_D_of_Dx_to_Q}--\ref{algoline:update:delete_from_Dtimes}).
Likewise, all non-pruned nodes in $\mD_{\supset}$ (note that pruning always considers all node collections $\Queue_{dup}$, $\Queue$, $\mD_{\checkmark}$, $\mD_{\times}$ and $\mD_{\supset}$) which are no longer supersets of any \emph{known} minimal diagnosis, are added to $\Queue$ again (lines \ref{algoline:update:process_Dsupset_start}--\ref{algoline:update:delete_from_Dsupset}).
Finally, those minimal diagnoses returned in the previous iteration 
and consistent with the latest added measurement (the elements of $\mD_{\checkmark}$), are put back to the ordered queue $\Queue$ (lines~\ref{algoline:update:process_Dcheckmark_start}--\ref{algoline:update:process_Dcheckmark_end}). 
This is necessary to preserve the best-first property, as there might be ``new'' minimal diagnoses for the current DPI that are more probable than known ones.\vspace{3pt} 

\textsc{prune}: Using its given argument $X$, the tree pruning runs through all (active and duplicate) nodes of the current tree (node collections $\Queue_{dup}$, $\Queue$, $\mD_{\supset}$ 
and $\mD_{calc}$ 
for call in line~\ref{algoline:dlabel:prune}, and $\Queue_{dup}$, $\Queue$, $\mD_{\supset}$, $\mD_{\times}$ and $\mD_{\checkmark}$ for call in line~\ref{algoline:update:prune}), and 
\begin{itemize}[noitemsep,leftmargin=*]
	\item \emph{(relabeling of old conflicts)} replaces by $X$ all labels $\mathsf{nd.cs}[i]$ which are proper supersets of $X$ for all nodes $\mathsf{nd}$ and for all $i = 1,\dots,|\mathsf{nd.cs}|$, and
	\item \emph{(deletion of redundant nodes)} deletes each redundant node $\mathsf{nd}$ for which $X$ is a witness of redundancy, and 
	\item \emph{(potential replacement of deleted nodes)} for each of the deleted nodes $\mathsf{nd}$, if available, uses a suitable (non-redundant) node $\mathsf{nd}'$ (constructed) from the elements of $\Queue_{dup}$ to replace $\mathsf{nd}$ by $\mathsf{nd}'$.
\end{itemize}
A node $\mathsf{nd}'$ qualifies as a \emph{replacement node for $\mathsf{nd}$} iff $\mathsf{nd}'$ is non-redundant and
$\mathsf{nd}'$ is \emph{set}-equal 
to $\mathsf{nd}$ (i.e., the \emph{sets}, not lists, of edge labels are equal).
This node replacement is necessary from the point of view of completeness (cf. \cite{greiner1989correction}).
Importantly, duplicates ($\Queue_{dup}$) must be pruned prior to all other nodes ($\Queue$, $\mD_{\supset}$, $\mD_{\times}$, $\mD_{\checkmark}$, $\mD_{calc}$), to guarantee that all surviving nodes in $\Queue_{dup}$ represent possible \emph{non-redundant} replacement nodes at the time other nodes are pruned. 

Additionally, the argument $X$ is used to update the conflicts stored for reuse (set $\mC_{calc}$), i.e., all proper supersets of $X$ are removed from $\mC_{calc}$ and $X$ is added to $\mC_{calc}$.

\begin{figure}[t]
	\centering
	\begin{minipage}[c]{\columnwidth}
		\xygraph{
			!{<0cm,0cm>;<1cm,0cm>:<0cm,-0.85cm>::}
			!{(-2.1,0) }*+{ \textcircled{\scriptsize 1}\langle1,2\rangle^{C} }="n1"
			!{(-4.2,.5) }*+{ \textcircled{\scriptsize 2}\langle2,3,4\rangle^{C} }="n2"
			"n1":"n2"_{1}
			!{(0,.5) }*+{ \textcircled{\scriptsize 3}\langle1,3,5\rangle^{C} }="n3"
			"n1":"n3"^{2}
			!{(-5.5,1.5) }*+{ \textcircled{\scriptsize 5}\langle3,4,5\rangle^{C} }="n4"
			"n2":"n4"_(0.6){2}
			!{(-3.9,1.5) }*+{ \textcircled{\scriptsize 6}\checkmark_{(\md_1)}^* }="n5"
			"n2":"n5"_{3}
			!{(-2.5,1.5) }*+{ \textcircled{\scriptsize 7}\checkmark_{(\md_2)}^* }="n6"
			"n2":"n6"^{4}
			!{(-1.2,1.5) }*+{ \textcircled{\scriptsize 4} dup }="n7"
			"n3":"n7"_(0.6){1}
			!{(0.1,1.5) }*+{ \textcircled{\scriptsize 8}\checkmark_{(\md_3)}^* }="n8"
			"n3":"n8"_{3}
			!{(1.5,1.5) }*+{ \textcircled{\scriptsize 9}\checkmark_{(\md_4)}^* }="n9"
			"n3":"n9"^{5}
			!{(-5.7,2.6) }*+{ \textcircled{\tiny 10} \XSup{\md_1} }="n10"
			"n4":"n10"_{3}
			!{(-3.9,2.6) }*+{ \textcircled{\tiny 11} \XSup{\md_2} }="n11"
			"n4":"n11"_{4}
			!{(-2.1,2.6) }*+{ \textcircled{\tiny 12} \XSup{\md_4} }="n12"
			"n4":"n12"^(0.6){5}
		}
		\vspace{2pt}
		\begin{center}
			\small 
			Iteration 1 ($\dpi_0$) $\quad\Rightarrow m_1:\;\, A\to C\phantom{\lnot}\;$ to $\Tn'$
		\end{center}
		\vspace{-5pt}
		
		\hdashrule[0.5ex]{\columnwidth}{1pt}{3mm} 
	\end{minipage}

	\begin{minipage}[c]{\columnwidth}
		\xygraph{
			!{<0cm,0cm>;<1cm,0cm>:<0cm,-0.85cm>::}
			!{(-2.1,0) }*+{ \textcircled{\scriptsize 1}\langle 1,2\rangle }="n13"
			!{(-4.2,0.5) }*+{ \textcircled{\scriptsize 2'}\langle 2,4\rangle }="n14"
			"n13":"n14"_{1}
			!{(0,0.5) }*+{ \textcircled{\scriptsize 3'}\langle 1,5\rangle }="n15"
			"n13":"n15"^{2}
			!{(-5,1.5) }*+{ \textcircled{\scriptsize 5}\langle 3,4,5\rangle }="n16"
			"n14":"n16"_{2}
			!{(-3.2,1.5) }*+{ \textcircled{\scriptsize 7}\checkmark_{(\md_2)} }="n17"
			"n14":"n17"^{4}
			!{(-1,1.5) }*+{ \textcircled{\scriptsize 4} dup }="n18"
			"n15":"n18"_{1}
			!{(0.8,1.5) }*+{ \textcircled{\scriptsize 9} \checkmark_{(\md_4)} }="n19"
			"n15":"n19"^{5}
			!{(-5.5,2.6) }*+{ \textcircled{\tiny 13} \langle 4,5\rangle^C }="n19"
			"n16":"n19"_{3}
			!{(-3.7,2.6) }*+{ \textcircled{\tiny 11} \XSup{\md_2} }="n20"
			"n16":"n20"_(0.4){4}
			!{(-1.9,2.6) }*+{ \textcircled{\tiny 12} \XSup{\md_4} }="n21"
			"n16":"n21"^(0.6){5}
			!{(-5.5,3.7) }*+{ \textcircled{\tiny 14} \XSup{\md_2} }="n22"
			"n19":"n22"_{4}
			!{(-3.7,3.7) }*+{ \textcircled{\tiny 15} \XSup{\md_4} }="n23"
			"n19":"n23"^(0.6){5}
		}
		\vspace{2pt}
		\begin{center}
			\small 
			Iteration 2 ($\dpi_1$) $\quad\Rightarrow m_2:\;\, A\to\lnot B\phantom{\lnot}\;$ to $\Tn'$
		\end{center}
		\vspace{-5pt}
		
		\hdashrule[0.5ex]{\columnwidth}{1pt}{3mm} 
	\end{minipage}

	\begin{minipage}[c]{\columnwidth}
		\xygraph{
			!{<0cm,0cm>;<1cm,0cm>:<0cm,-0.85cm>::}
			!{(-2.1,0) }*+{ \textcircled{\scriptsize 1'}\langle1\rangle }="n24"
			!{(-4.2,.5) }*+{ \textcircled{\scriptsize 2'}\langle2,4\rangle }="n25"
			"n24":"n25"_{1}
			!{(-5,1.5) }*+{ \textcircled{\scriptsize 5}\langle3,4,5\rangle }="n26"
			"n25":"n26"_{2}
			!{(-3.2,1.5) }*+{ \textcircled{\scriptsize 7}\checkmark_{(\md_2)} }="n27"
			"n25":"n27"^{4}
			!{(-5.5,2.6) }*+{ \textcircled{\tiny 13}\langle4,5\rangle }="n28"
			"n26":"n28"_{3}
			!{(-3.7,2.6) }*+{ \textcircled{\tiny 11}\XSup{\md_2} }="n29"
			"n26":"n29"_{4}
			!{(-1.9,2.6) }*+{ \textcircled{\tiny 16}\langle3,4\rangle^C }="n30"
			"n26":"n30"^(0.6){5}
			!{(-5.5,3.7) }*+{ \textcircled{\tiny 14}\XSup{\md_2} }="n31"
			"n28":"n31"_{4}
			!{(-3.7,3.7) }*+{ \textcircled{\tiny 18}\checkmark_{(\md_5)}^* }="n32"
			"n28":"n32"_{5}
			!{(-1.9,3.7) }*+{ \textcircled{\tiny 17} dup }="n33"
			"n30":"n33"_{3}
			!{(-0.1,3.7) }*+{ \textcircled{\tiny 19}\XSup{\md_2} }="n34"
			"n30":"n34"_{4}
		}
		\vspace{2pt}
		\begin{center}
			\small 
			Iteration 3 ($\dpi_2$) $\quad\Rightarrow m_3:\;\, A\to\lnot C\phantom{\lnot}\;$ to $\Tp'$
		\end{center}
		\vspace{-5pt}
		
		\hdashrule[0.5ex]{\columnwidth}{1pt}{3mm}
	\end{minipage}
	
	\begin{minipage}[c]{\columnwidth}
		\xygraph{
			!{<0cm,0cm>;<1cm,0cm>:<0cm,-0.85cm>::}
			!{(-0.3,0) }*+{ \textcircled{\scriptsize 1'}\langle1\rangle }="n35"
			!{(-2.3,.5) }*+{ \textcircled{\scriptsize 2''}\langle4\rangle }="n36"
			"n35":"n36"_{1}
			!{(-1.3,1.5) }*+{ \textcircled{\scriptsize 7}\checkmark_{(\md_2)} }="n37"
			"n36":"n37"^{4}
			!{(-4.3,1.5) }*+{ \hphantom{3pt}}="n38"
		}
		\vspace{2pt}
		\begin{center}
			\small 
			Iteration 4 ($\dpi_3$)
		\end{center}
		\vspace{-5pt} 
		
		\hdashrule[0.5ex]{\columnwidth}{1pt}{3mm} 
	\end{minipage}

	\caption{DynamicHS executed on example DPI given in Tab.~\ref{tab:example_DPI}.}
	\label{fig:dynhs_example}
\end{figure}




\begin{figure}[t]
	\centering
	\begin{minipage}[c]{\columnwidth}
		\xygraph{
			!{<0cm,0cm>;<1cm,0cm>:<0cm,-0.85cm>::}
			!{(-2.1,0) }*+{ \textcircled{\scriptsize 1}\langle1,2\rangle^{C} }="n1"
			!{(-4.2,.5) }*+{ \textcircled{\scriptsize 2}\langle2,3,4\rangle^{C} }="n2"
			"n1":"n2"_{1}
			!{(0,.5) }*+{ \textcircled{\scriptsize 3}\langle1,3,5\rangle^{C} }="n3"
			"n1":"n3"^{2}
			!{(-5.5,1.5) }*+{ \textcircled{\scriptsize 4}\langle3,4,5\rangle^{C} }="n4"
			"n2":"n4"_(0.6){2}
			!{(-3.9,1.5) }*+{ \textcircled{\scriptsize 5}\checkmark_{(\md_1)}^* }="n5"
			"n2":"n5"_{3}
			!{(-2.5,1.5) }*+{ \textcircled{\scriptsize 6}\checkmark_{(\md_2)}^* }="n6"
			"n2":"n6"^{4}
			!{(-1.2,1.5) }*+{ \textcircled{\scriptsize 7} \times }="n7"
			"n3":"n7"_(0.6){1}
			!{(0.1,1.5) }*+{ \textcircled{\scriptsize 8}\checkmark_{(\md_3)}^* }="n8"
			"n3":"n8"_{3}
			!{(1.5,1.5) }*+{ \textcircled{\scriptsize 9}\checkmark_{(\md_4)}^* }="n9"
			"n3":"n9"^{5}
			!{(-5.7,2.6) }*+{ \textcircled{\tiny 10} \XSup{\md_1} }="n10"
			"n4":"n10"_{3}
			!{(-3.9,2.6) }*+{ \textcircled{\tiny 11} \XSup{\md_2} }="n11"
			"n4":"n11"_{4}
			!{(-2.1,2.6) }*+{ \textcircled{\tiny 12} \XSup{\md_4} }="n12"
			"n4":"n12"^(0.6){5}
		}
		\vspace{2pt}
		\begin{center}
			\small 
			Iteration 1 ($\dpi_0$) $\quad\Rightarrow m_1:\;\, A\to C\phantom{\lnot}\;$ to $\Tn'$
		\end{center}
		\vspace{-5pt}
		
		\hdashrule[0.5ex]{\columnwidth}{1pt}{3mm} 
	\end{minipage}
	
	\begin{minipage}[c]{\columnwidth}
		\xygraph{
			!{<0cm,0cm>;<1cm,0cm>:<0cm,-0.85cm>::}
			!{(-2.1,0) }*+{ \textcircled{\tiny 13}\langle1,2\rangle^{C} }="n1"
			!{(-4.2,.5) }*+{ \textcircled{\tiny 14}\langle2,4\rangle^{C} }="n2"
			"n1":"n2"_{1}
			!{(0,.5) }*+{ \textcircled{\tiny 15}\langle1,5\rangle^{C} }="n3"
			"n1":"n3"^{2}
			!{(-5.5,1.5) }*+{ \textcircled{\tiny 16}\langle4,5\rangle^{C} }="n4"
			"n2":"n4"_(0.6){2}
			!{(-3,1.5) }*+{ \textcircled{\tiny 17}\checkmark_{(\md_2)}^* }="n6"
			"n2":"n6"^{4}
			!{(-0.7,1.5) }*+{ \textcircled{\tiny 18} \times }="n7"
			"n3":"n7"_(0.6){1}
			!{(1.5,1.5) }*+{ \textcircled{\tiny 19}\checkmark_{(\md_4)}^* }="n9"
			"n3":"n9"^{5}
			!{(-5.7,2.6) }*+{ \textcircled{\tiny 20} \XSup{\md_2} }="n11"
			"n4":"n11"_{4}
			!{(-3.5,2.6) }*+{ \textcircled{\tiny 21} \XSup{\md_4} }="n12"
			"n4":"n12"^(0.6){5}
		}
		\vspace{2pt}
		\begin{center}
			\small 
			Iteration 2 ($\dpi_1$) $\quad\Rightarrow m_2:\;\, A\to\lnot B\phantom{\lnot}\;$ to $\Tn'$
		\end{center}
		\vspace{-5pt}
		
		\hdashrule[0.5ex]{\columnwidth}{1pt}{3mm} 
	\end{minipage}
	
	\begin{minipage}[c]{\columnwidth}
		\xygraph{
			!{<0cm,0cm>;<1cm,0cm>:<0cm,-0.85cm>::}
			!{(-2.1,0) }*+{ \textcircled{\tiny 22}\langle1\rangle^{C} }="n1"
			!{(-4.2,.5) }*+{ \textcircled{\tiny 23}\langle2,4\rangle^{C} }="n2"
			"n1":"n2"_{1}
			!{(-5.5,1.5) }*+{ \textcircled{\tiny 24}\langle4,5\rangle^{C} }="n4"
			"n2":"n4"_(0.6){2}
			!{(-3,1.5) }*+{ \textcircled{\tiny 25}\checkmark_{(\md_2)}^* }="n6"
			"n2":"n6"^{4}
			!{(-5.7,2.6) }*+{ \textcircled{\tiny 26} \XSup{\md_2} }="n11"
			"n4":"n11"_{4}
			!{(-3.5,2.6) }*+{ \textcircled{\tiny 27} \langle3,4\rangle^{C} }="n12"
			"n4":"n12"^(0.6){5}
			!{(-4.3,3.7) }*+{ \textcircled{\tiny 28} \checkmark_{(\md_5)}^* }="n13"
			"n12":"n13"_{3}
			!{(-2.1,3.7) }*+{ \textcircled{\tiny 29} \XSup{\md_2} }="n14"
			"n12":"n14"^{4}
		}
		\vspace{2pt}
		\begin{center}
			\small 
			Iteration 3 ($\dpi_2$) $\quad\Rightarrow m_3:\;\, A\to\lnot C\phantom{\lnot}\;$ to $\Tp'$
		\end{center}
		\vspace{-5pt}
		
		\hdashrule[0.5ex]{\columnwidth}{1pt}{3mm} 
	\end{minipage}	
	
	\begin{minipage}[c]{\columnwidth}
		\xygraph{
			!{<0cm,0cm>;<1cm,0cm>:<0cm,-0.85cm>::}
			!{(-0.3,0) }*+{ \textcircled{\scriptsize 30}\langle1\rangle^C }="n35"
			!{(-2.3,.5) }*+{ \textcircled{\scriptsize 31}\langle4\rangle^C }="n36"
			"n35":"n36"_{1}
			!{(-1.3,1.5) }*+{ \textcircled{\scriptsize 32}\checkmark_{(\md_2)}^* }="n37"
			"n36":"n37"^{4}
			!{(-4.3,1.5) }*+{ \hphantom{3pt}}="n38"
		}
		\vspace{2pt}
		\begin{center}
			\small 
			Iteration 4 ($\dpi_3$)
		\end{center}
		\vspace{-5pt} 
		
		\hdashrule[0.5ex]{\columnwidth}{1pt}{3mm} 
	\end{minipage}

	\caption{HS-Tree executed on example DPI given in Tab.~\ref{tab:example_DPI}.}
	\label{fig:hs_example}
\end{figure}

\begin{table}
	\scriptsize
	\centering
	\begin{tabular}{@{}c@{\,}|@{\,}c@{\,}|@{\,}c@{\,}|@{\,}c@{\,}|@{\,}c@{}}
		iteration $j$ & $\Tp'$ & $\Tn'$ & $\sol(\dpi_{j-1})$ & $\conf(\dpi_{j-1})$ \\
		\hline
		\multirow{2}{*}{1}         & \multirow{2}{*}{--}   & \multirow{2}{*}{--}   &  	$[1,3],[1,4],$      & 	$\tuple{1,2},\tuple{2,3,4},$           \\ 
		& & & $[2,3],[2,5]$ & $\tuple{1,3,5},\tuple{3,4,5}$ \\ \hline
		\multirow{2}{*}{2}  		  & \multirow{2}{*}{--}   &  \multirow{2}{*}{$A\to C$}  &  \multirow{2}{*}{$[1,4],[2,5]$}         &   $\tuple{1,2},\tuple{2,4},$        \\
		&&&& $\tuple{1,5},\tuple{4,5}$   \\ \hline
		\multirow{2}{*}{3}         & \multirow{2}{*}{--}   &  \multirow{2}{*}{$A\to C, \; A\to \lnot B$}  &   \multirow{2}{*}{$[1,4],[1,2,3,5]$}        &   $\tuple{1},\tuple{2,4},$        \\
		&&&& $\fbox{\tuple{3,4}},\tuple{4,5}$ \\ \hline
		4         &  $A\to \lnot C$  &  $A\to C, \; A\to \lnot B$  &  $[1,4]$         &  $\tuple{1},\tuple{4}$        
	\end{tabular}
	\caption{\small Evolution of minimal diagnoses and minimal conflicts after successive extension of the example DPI $\dpi_0$ (Tab.~\ref{tab:example_DPI}) by positive ($\Tp'$) or negative ($\Tn'$) measurements $m_i$ shown in Figures~\ref{fig:dynhs_example} and \ref{fig:hs_example}.
		Newly arisen conflicts (cf.\ $\mc_{\mathit{new}}$ in Property~\ref{property:impact_of_DPI_transition}.\ref{property:impact_of_DPI_transition:enum:some_conflict_shrinks_or_new_conflict_arises}) 
		are framed.}
	\label{tab:diag_conf_evolution_example}
\end{table}

\begin{example}\label{ex:algo_description}
Consider the propositional DPI $\dpi_0$ in Tab.~\ref{tab:example_DPI}. The goal is 
to locate
the faulty axioms in the KB $\mo$ 
that prevent the satisfaction of given measurements $\Tp$ and $\Tn$ (here, only one negative measurement $\lnot A$ is given, i.e., $\lnot A$ must not be entailed by the correct KB). We now illustrate the workings of HS-Tree (Fig.~\ref{fig:hs_example}) and DynamicHS (Fig.~\ref{fig:dynhs_example}) in terms of a complete sequential diagnosis session for $\dpi_0$ under the assumption that $[1,4]$ is the actual diagnosis. The initial 
set of minimal conflicts and diagnoses can be read from Tab.~\ref{tab:diag_conf_evolution_example}. 
\vspace{1.5pt} 

\noindent\emph{Inputs (Sequential Diagnosis):} 
We set $\ld := 5$ (if existent, compute five diagnoses per iteration), $\mathsf{heur}$ to be the ``split-in-half'' heuristic \cite{Shchekotykhin2012} (prefers measurements the more, the more diagnoses they eliminate in the worst case), and $\pr$ in a way 
the hitting set trees are constructed breadth-first.\vspace{1.5pt} 

\noindent\emph{Notation in Figures:} Axioms 
$\tax_i$ are simply referred to by $i$ (in node and edge labels).
Numbers $\textcircled{\scriptsize i}$ indicate the chronological 
node labeling
order.
%
We tag conflicts $\tuple{1,\dots,k}$ by $^C$ if they are freshly computed (\textsc{findMinConflict} call, line~\ref{algoline:dlabel:qx_2}, \textsc{dLabel}), and leave them 
untagged
if
they result from a redundancy check and subsequent relabeling
(lines~\ref{algoline:update:call_redundant_function}--\ref{algoline:update:prune}, \textsc{updateTree}).
For the leaf nodes, we use the following labels:
$\checkmark_{(\md_i)}$ for a minimal diagnosis, named $\md_i$, stored in $\mD_{calc}$;
$\times$ for a duplicate in HS-Tree (see \ref{enum:hstree:label:duplicate} criterion);
$\times_{(\supset \md_i)}$ for a non-minimal diagnosis (stored in $\mD_{\supset}$ by DynamicHS), where $\md_i$ is one diagnosis that proves the non-minimality; and
$\mathit{dup}$ for a duplicate in DynamicHS (stored in $\Queue_{dup}$).
Branches representing minimal diagnoses are additionally tagged by a $^*$ if logical reasoning (\textsc{findMinConflict} call, line~\ref{algoline:dlabel:qx_2}, \textsc{dLabel} function) is necessary to prove it is a diagnosis, and untagged otherwise 
(i.e., branch is diagnosis from previous iteration, stored in $\mD_{\checkmark}$; only pertinent to DynamicHS).
     

\noindent\emph{Iteration~1:} 
In the first iteration, 
HS-Tree and DynamicHS essentially 
build the same tree (compare 
Figs.~\ref{fig:dynhs_example} and \ref{fig:hs_example}). The only difference is that DynamicHS stores the duplicates 
and non-minimal diagnoses (labels $\mathit{dup}$ and $\times_{(\supset \md_i)}$),
whereas HS-Tree discards them (labels $\times$ and $\times_{(\supset \md_i)}$). Note, duplicates are stored by DynamicHS at \emph{generation} time (line~\ref{algoline:dyn:add_to_Qdup}), hence the found duplicate ($\mathit{dup}$) has number \textcircled{\scriptsize 4} (not \textcircled{\scriptsize 7}, as in HS-Tree). The diagnoses computed by both algorithms are $\{\md_1,\md_2,\md_3,\md_4\} = \{[1,3],[1,4],[2,3],[2,5]\}$ (cf.\ Tab.~\ref{tab:diag_conf_evolution_example}).
Notably, the returned diagnoses \emph{must} be equal for both algorithms in any iteration (given same parameters $\ld$ and $\pr$) since both are sound, complete and best-first minimal diagnosis searches. Thus, when using the same measurement selection (and heuristic $\mathsf{heur}$), both methods \emph{must} also give rise to the same proposed next measurement $m_i$ in each iteration.\vspace{1.5pt} 

\noindent\emph{First Measurement:} Accordingly, both algorithms lead to the first measurement $m_1: A \to C$, 
which corresponds to the question ``Does $A$ imply $C$?''. This measurement
turns out to be negative, e.g., by consulting an expert for the domain described by $\mo$, and is therefore added to $\Tn'$. This effectuates a transition from 
$\dpi_0$ to a new DPI $\dpi_1$ (which includes the additional element $A \to C$ in $\Tn'$), and thus a change of the relevant minimal diagnoses and conflicts (see Tab.~\ref{tab:diag_conf_evolution_example}). \vspace{1.5pt}  

\noindent\emph{Tree Update:} Starting from the second iteration ($\dpi_1$), HS-Tree and DynamicHS behave differently. Whereas the former constructs a new tree from scratch for $\dpi_1$, DynamicHS 
runs \textsc{updateTree}
to make the existing tree (built for $\dpi_0$) reusable for $\dpi_1$. In the course of this tree update, two witnesses of redundancy (minimal conflicts $\tuple{2,4}$, $\tuple{1,5}$) are found while analyzing the (conflicts along the) branches of the two invalidated diagnoses $[1,3]$ and $[2,3]$ (\textcircled{\scriptsize 6} and \textcircled{\scriptsize 8}). 
E.g., $\mathsf{nd}=[1,3]$ is redundant since the 
conflict $\mathsf{nd.cs}[2] = \tuple{2,3,4}$ is a proper superset of the \emph{now minimal} conflict $X = \tuple{2,4}$ \emph{and} $\mathsf{nd}$'s outgoing edge of $\mathsf{nd.cs}[2]$ is $\mathsf{nd}[2] = 3$ which is an element of $\mathsf{nd.cs}[2]\setminus X = \setof{3}$.  
Since 
stored duplicates (here: only $[2,1]$) do not allow the construction of a replacement node for any of the redundant branches $[1,3]$ and $[2,3]$, both are removed from the tree. Further, 
the witnesses of redundancy replace the non-minimal conflicts at 
\textcircled{\scriptsize 2} and \textcircled{\scriptsize 3}, which is signified by the new numbers \textcircled{\scriptsize 2'} and \textcircled{\scriptsize 3'}.
%

Other than that, only a single further change is induced by \textsc{updateTree}. Namely, the branch $[1,2,3]$, a non-minimal diagnosis for $\dpi_0$, is returned to 
$\Queue$ (unlabeled nodes) because there is no longer a diagnosis in the tree witnessing its non-minimality (both such witnesses $[1,3]$ and $[2,3]$ have been discarded). Note that, first, $[1,2,3]$ is in fact no longer a hitting set of all minimal conflicts for $\dpi_1$ (cf.\ Tab.~\ref{tab:diag_conf_evolution_example}) and, second, there is still a non-minimality witness for all other branches (\textcircled{\tiny 12} and \textcircled{\tiny 13}) representing non-minimal diagnoses for $\dpi_0$, which is why they remain labeled by $\times_{(\supset \md_i)}$. \vspace{1.5pt} 

\noindent\emph{Iteration~2:} Observe the crucial differences between HS-Tree and DynamicHS in the second iteration (cf.\ Figs.~\ref{fig:dynhs_example} and \ref{fig:hs_example}). 

First, while HS-Tree has to compute all conflicts that label nodes by (potentially expensive) \textsc{findMinConflict} calls ($^C$ tags), DynamicHS has
(cheaply) reduced existing conflicts
during pruning (see above).
Note, however, not all conflicts are necessarily always kept up-to-date after a DPI-transition (\emph{lazy updating policy}). 
E.g., node \textcircled{\scriptsize 5} is still labeled by the non-minimal conflict $\tuple{3,4,5}$ after \textsc{updateTree} terminates. Hence, the subtree comprising nodes \textcircled{\tiny 13}, \textcircled{\tiny 14} and \textcircled{\tiny 15} is not present in HS-Tree. 
Importantly, this lazy updating does not counteract sound- or completeness of DynamicHS, and the overhead incurred by such additional nodes is generally minor (all these nodes must be non-minimal diagnoses and are thus not further expanded).
%
Second, the verification of the minimal diagnoses ($\md_2$, $\md_4$) found in iteration~2 requires logical reasoning in HS-Tree (see $^*$ tags of $\checkmark$ nodes) whereas it comes for free in DynamicHS (storage and reuse of $\mD_{\checkmark}$).\vspace{1.5pt}

\noindent\emph{Remaining Execution:} After the second measurement $m_2$ is added to $\Tn'$, causing a DPI-transition once again, DynamicHS reduces the conflict that labels the root node. This leads to the pruning of the complete right subtree. The left subtree is then further expanded in iteration~3 (see generated nodes \textcircled{\tiny 16}, \textcircled{\tiny 17}, \textcircled{\tiny 18} and \textcircled{\tiny 19}) until the two leading diagnoses $\md_2 = [1,4]$ and $\md_5 = [1,2,3,5]$ are located and the queue of unlabeled nodes becomes empty (which proves that no further minimal diagnoses exist). Eventually, the addition of the third measurement $m_3$ to $\Tp'$ brings sufficient information to isolate the actual diagnosis. This is reflected by a pruning of all branches except for the one representing the actual diagnosis $[1,4]$.\vspace{1.5pt}

\noindent\emph{Comparison (expensive operations):} Generally, calling \textsc{findMinConflict} (FC) is (clearly) more costly than \textsc{redundant} (RD), which in turn is more costly than a single consistency check (CC). HS-Tree requires 14/0/9, DynamicHS only 6/4/5 FC/RD/CC calls. This reduction of costly reasoning is the crucial advantage of DynamicHS over HS-Tree.\qed
\end{example}

\noindent\textbf{DynamicHS: Properties.} A proof of the following correctness theorem for DynamicHS can be found in \cite[Sec.~12.4]{Rodler2015phd}:
\begin{theorem}
DynamicHS is a sound, complete and best-first (as per $\pr$) minimal diagnosis computation method.
\end{theorem}

\section{First Experiment Results}
\label{sec:eval}
We provide a quick glance at first results of ongoing evaluations, where we compare 
HS-Tree and DynamicHS when applied for fault localization in (inconsistent) real-world KBs (same dataset as used in \cite[Tabs.~8+12]{Shchekotykhin2012}). In this domain, HS-Tree is the most often used diagnosis search method. Average savings of DynamicHS over HS-Tree (both using same $\ld:=6$ and random $\pr$) wrt.\ \emph{(i)} \emph{number of expensive reasoner calls} (\textsc{findMinConflict}), and \emph{(ii)} \emph{running time}, amounted to (i) 59\,\% and (ii) 42\,\%. Notably, DynamicHS required less time in \emph{all} observed sequential sessions.

\section{Related Work}
Diagnosis algorithms can be compared along multiple axes, e.g., best-first vs.\ any-first, complete vs.\ incomplete, stateful vs.\ stateless, black-box (reasoner used as oracle) vs.\ glass-box (reasoner modified), and on-the-fly vs. preliminary (conflict computation). 
DynamicHS is best-first, complete, stateful, black-box, and on-the-fly. The most similar algorithms in terms of these features are: \emph{(i)~StaticHS} \cite{DBLP:conf/socs/RodlerH18}: same features, but different focus, which is on reducing measurement conduction time (cf.\ Sec.~\ref{sec:intro}); \emph{(ii)~Inv-HS-Tree} \cite{Shchekotykhin2014}: same features, but any-first, can handle problems with high-cardinality diagnoses that HS-Tree (and DynamicHS) cannot; \emph{(iii)~GDE and its successors} \cite{dekleer1987}: same features, but not black-box (e.g., uses bookkeeping of justifications for computed entailments).
Generally, depending on the specific application 
setting 
of a diagnosis algorithm, different features are (un)necessary or (un)desirable. E.g., while HS-Tree-based tools are rather successful in the domain of knowledge-based systems, GDE-based ones prevail in the circuit domain 
(perhaps since bookkeeping methods are better suited for the entailment-justification structure of hardware than of KB systems\footnote{Based on a discussion in plenum at DX'17 in Brescia.}).

\section{Conclusions}
We suggest a novel sound, complete and best-first diagnosis computation method for sequential diagnosis based on Reiter's HS-Tree which aims at reducing expensive reasoning by the maintenance of a search data structure throughout a diagnostic session. First experimental results are very promising and attest the reasonability of the approach.
\\

\noindent\textbf{Acknowledgments:} Thanks to Manuel Herold for implementing DynamicHS and the experiments. 


%
%


\begin{thebibliography}{10}
	
	\bibitem{abreu2011simultaneous}
	R~Abreu, P~Zoeteweij, and AJ Van~Gemund.
	\newblock Simultaneous debugging of software faults.
	\newblock {\em J. of Syst. and Softw.}, 84(4):573--586, 2011.
	
	\bibitem{dekleer1987}
	J~de~Kleer and BC Williams.
	\newblock {Diagnosing multiple faults}.
	\newblock {\em Artif. Intell.}, 32(1):97--130, 1987.
	
	\bibitem{DBLP:journals/ai/FelfernigFJS04}
	A~Felfernig, G~Friedrich, D~Jannach, and M~Stumptner.
	\newblock {Consistency-based diagnosis of configuration knowledge bases}.
	\newblock {\em Artif. Intell.}, 152(2):213--234, 2004.
	
	\bibitem{felfernig2008intelligent}
	A~Felfernig, E~Teppan, G~Friedrich, and K~Isak.
	\newblock Intelligent debugging and repair of utility constraint sets in
	knowledge-based recommender applications.
	\newblock In {\em 13th Int'l Conf. on
		Intell. User Interf.}, pp. 217--226, 2008.
	
	\bibitem{friedrich2005gdm}
	G~Friedrich and K~Schekotihin.
	\newblock {A General Diagnosis Method for Ontologies}.
	\newblock In {\em ISWC'05}, pp. 232--246, 2005.
	
	\bibitem{friedrich1999model}
	G~Friedrich, M~Stumptner, and F~Wotawa.
	\newblock Model-based diagnosis of hardware designs.
	\newblock {\em Artif. Intell.}, 111(1-2):3--39, 1999.
	
	\bibitem{grau2008owl}
	BC Grau, I~Horrocks, B~Motik, B~Parsia, P~Patel-Schneider, and
	U~Sattler.
	\newblock {OWL 2}: The next step for {OWL}.
	\newblock {\em JWS}, 6(4):309--322, 2008.
	
	\bibitem{greiner1989correction}
	R~Greiner, BA Smith, and RW Wilkerson.
	\newblock {A correction to the algorithm in Reiter's theory of diagnosis}.
	\newblock {\em Artif. Intell.}, 41(1):79--88, 1989.
	
	\bibitem{Horridge2011a}
	M~Horridge.
	\newblock {\em Justification based Explanation in Ontologies}.
	\newblock PhD thesis, Univ. of Manchester, 2011.
	
	\bibitem{junker04}
	U~Junker.
	\newblock {QUICKXPLAIN: Preferred Explanations and Relaxations for
		Over-Constrained Problems}.
	\newblock In {\em AAAI'04}, pp. 167--172, 2004.
	
	\bibitem{Kalyanpur2006a}
	A~Kalyanpur.
	\newblock {\em {Debugging and Repair of OWL Ontologies}}.
	\newblock PhD thesis, Univ. of Maryland, 2006.
	
	\bibitem{Kleer1992}
	J~de Kleer, AK Mackworth, and R~Reiter.
	\newblock {Characterizing diagnoses and systems}.
	\newblock {\em Artif. Intell.}, 56, 1992.
	
	\bibitem{meilicke2011}
	C~Meilicke.
	\newblock {\em {Alignment Incoherence in Ontology Matching}}.
	\newblock PhD thesis, Univ. Mannheim, 2011.
	
	\bibitem{pattipati1990}
	KR Pattipati and MG Alexandridis.
	\newblock Application of heuristic search and information theory to sequential
	fault diagnosis.
	\newblock {\em IEEE Transactions on Systems, Man, and Cybernetics},
	20(4):872--887, 1990.
	
	\bibitem{Reiter87}
	R~Reiter.
	\newblock {A Theory of Diagnosis from First Principles}.
	\newblock {\em Artif. Intell.}, 32(1):57--95, 1987.
	
	\bibitem{Rodler2015phd}
	P~Rodler.
	\newblock {\em {Interactive Debugging of Knowledge Bases}}.
	\newblock PhD thesis, Univ. Klagenfurt, 2015.
	\newblock {\em CoRR}, abs/1605.05950v1
	
	\bibitem{rodler17dx_activelearning}
	P~Rodler.
	\newblock On active learning strategies for sequential diagnosis.
	\newblock In {\em DX'17}, pp. 264--283, 2018.
	
	\bibitem{DBLP:conf/ieaaie/RodlerE19}
	P~Rodler and M~Eichholzer.
	\newblock On the usefulness of different expert question types for fault localization in ontologies.
	\newblock In {\em IEA/AIE'19}, pp. 360--375, 2019.
	
	\bibitem{DBLP:conf/socs/RodlerH18}
	P~Rodler and M~Herold.
	\newblock Statichs: {A} variant of reiter's hitting set tree for efficient
	sequential diagnosis.
	\newblock In {\em SOCS'18},
	pp. 72--80, 2018.
	
	\bibitem{DBLP:journals/kbs/RodlerJSF19}
	P~Rodler, D~Jannach, K~Schekotihin, and P~Fleiss.
	\newblock Are query-based ontology debuggers really helping knowledge engineers?
	\newblock {\em Knowl.-Based Syst.}, 179:92--107, 2019.
	
	\bibitem{rodler17dx_reducing}
	P~Rodler and K~Schekotihin.
	\newblock Reducing model-based diagnosis to KB debugging.
	\newblock In {\em DX'17}, pp. 284--296, 2018.
	
	\bibitem{DBLP:conf/ruleml/RodlerS18}
	P~Rodler and W~Schmid.
	\newblock On the impact and proper use of heuristics in test-driven ontology
	debugging.
	\newblock In {\em RuleML+RR'18}, pp. 164--184, 2018.
	
	\bibitem{DBLP:journals/corr/Rodler2017}
	P~Rodler, W~Schmid, and K~Schekotihin.
	\newblock A generally applicable, highly scalable measurement computation and
	optimization approach to sequential model-based diagnosis.
	\newblock {\em CoRR}, abs/1711.05508, 2017.
	
	\bibitem{Rodler2013}
	P~Rodler, K~Shchekotykhin, P~Fleiss, and G~Friedrich.
	\newblock {RIO: Minimizing User Interaction in Ontology Debugging}.
	\newblock In {\em RR'13}, pp. 153--167, 2013.
	
	\bibitem{Shchekotykhin2012}
	K~Shchekotykhin, G~Friedrich, P~Fleiss, and P~Rodler.
	\newblock {Interactive Ontology Debugging: Two Query Strategies for Efficient
		Fault Localization}.
	\newblock {\em JWS}, 12-13:88--103, 2012.
	
	\bibitem{Shchekotykhin2014}
	K~Shchekotykhin, G~Friedrich, P~Rodler, and P~Fleiss.
	\newblock {Sequential diagnosis of high cardinality faults in KBs
		by direct diagnosis generation}.
	\newblock In {\em ECAI'14}, pp. 813--818, 2014.
	
	\bibitem{wotawa2010fault}
	F~Wotawa.
	\newblock Fault localization based on dynamic slicing and hitting-set
	computation.
	\newblock In {\em 10th Int'l Conf. on Quality Softw.},
	pp. 161--170, 2010.
	
	
\end{thebibliography}
 
 \fontsize { 9.5pt }{ 10.5pt } 
 \selectfont

%
%
%
%
%
\end{document}